\documentclass[11pt,a4paper]{article}
\usepackage[hyperref]{emnlp2018}
\usepackage{times}
\usepackage{latexsym}
\usepackage{booktabs}
\usepackage{breqn}
\usepackage{natbib}
\usepackage{tikz}
\usepackage{todonotes}
\usepackage{subcaption}
\usetikzlibrary{shapes.geometric,arrows,chains,matrix,positioning,scopes,calc}
\tikzstyle{mybox} = [draw=white, rectangle]
\usepackage{url}
\usepackage{flushend}
\usepackage[title=normal,subtle]{savetrees}

\def \Figname {Figure}
\newcommand{\Figref}[1]{\Figname~\ref{#1}}

\def \tabname {Table}
\newcommand{\tabref}[1]{\tabname~\ref{#1}}

\aclfinalcopy 

\def \eg {e.g., }
\def \ie {i.e., }

\definecolor{camlightblue}{rgb}{0.601 , 0.8, 1}
\definecolor{camdarkblue}{rgb}{0, 0.203, 0.402}
\definecolor{camred}{rgb}{1, 0.203, 0}
\definecolor{camyellow}{rgb}{1, 0.8, 0}
\definecolor{lightblue}{rgb}{0, 0, 0.80}
\definecolor{white}{rgb}{1, 1, 1}
\definecolor{whiteblue}{rgb}{0.80, 0.80, 1}
\definecolor{nice-red}{HTML}{E41A1C}
\definecolor{nice-orange}{HTML}{FF7F00}
\definecolor{nice-yellow}{HTML}{FFC020}
\definecolor{nice-green}{HTML}{4DAF4A}
\definecolor{nice-blue}{HTML}{377EB8}
\definecolor{nice-purple}{HTML}{984EA3}

\makeatletter
\let\UrlSpecialsOld\UrlSpecials
\def\UrlSpecials{\UrlSpecialsOld\do\/{\Url@slash}\do\_{\Url@underscore}}%
\def\Url@slash{\@ifnextchar/{\kern-.11em\mathchar47\kern-.2em}%
    {\kern-.0em\mathchar47\kern-.08em\penalty\UrlBigBreakPenalty}}
\makeatother

\title{Prior Attention for \\ Style-aware Sequence-to-Sequence Models}

\author{Lucas Sterckx, Johannes Deleu, Chris Develder  \and Thomas Demeester\\
  IDLab, Ghent University - imec \\
  {\tt firstname.lastname@ugent.be} 
}
\date{}

\begin{document}
\maketitle
\begin{abstract}
We extend sequence-to-sequence models with the possibility to control the characteristics or style of the generated output, via attention that is generated a priori (before decoding) from a latent code vector.
After training an initial attention-based sequence-to-sequence model, we use a variational auto-encoder conditioned on representations of input sequences and a latent code vector space to generate attention matrices.
By sampling the code vector from specific regions of this latent space during decoding and imposing prior attention generated from it in the seq2seq model, output can be steered towards having certain attributes. 
This is demonstrated for the task of sentence simplification, where the latent code vector allows control over output length and lexical simplification, and enables fine-tuning to optimize for different evaluation metrics.
\end{abstract}

\section{Introduction}
\label{sec:intro}
Apart from its application to machine translation, the \textit{encoder-decoder} or \textit{sequence-to-sequence} (seq2seq) paradigm has been successfully applied to monolingual text-to-text tasks including simplification~\cite{P17-2014}, paraphrasing~\cite{mallinson-sennrich-lapata:2017:EACLlong}, style transfer~\cite{jhamtani-EtAl:2017:StyVa}, sarcasm interpretation~\cite{peled-reichart:2017:Long}, automated lyric annotation~\cite{sterckx-EtAl:2017:EMNLP2017}~and dialogue systems~\cite{serban2016building}.
\begin{figure*}[t]
\centering
\includegraphics[width=0.8\textwidth]{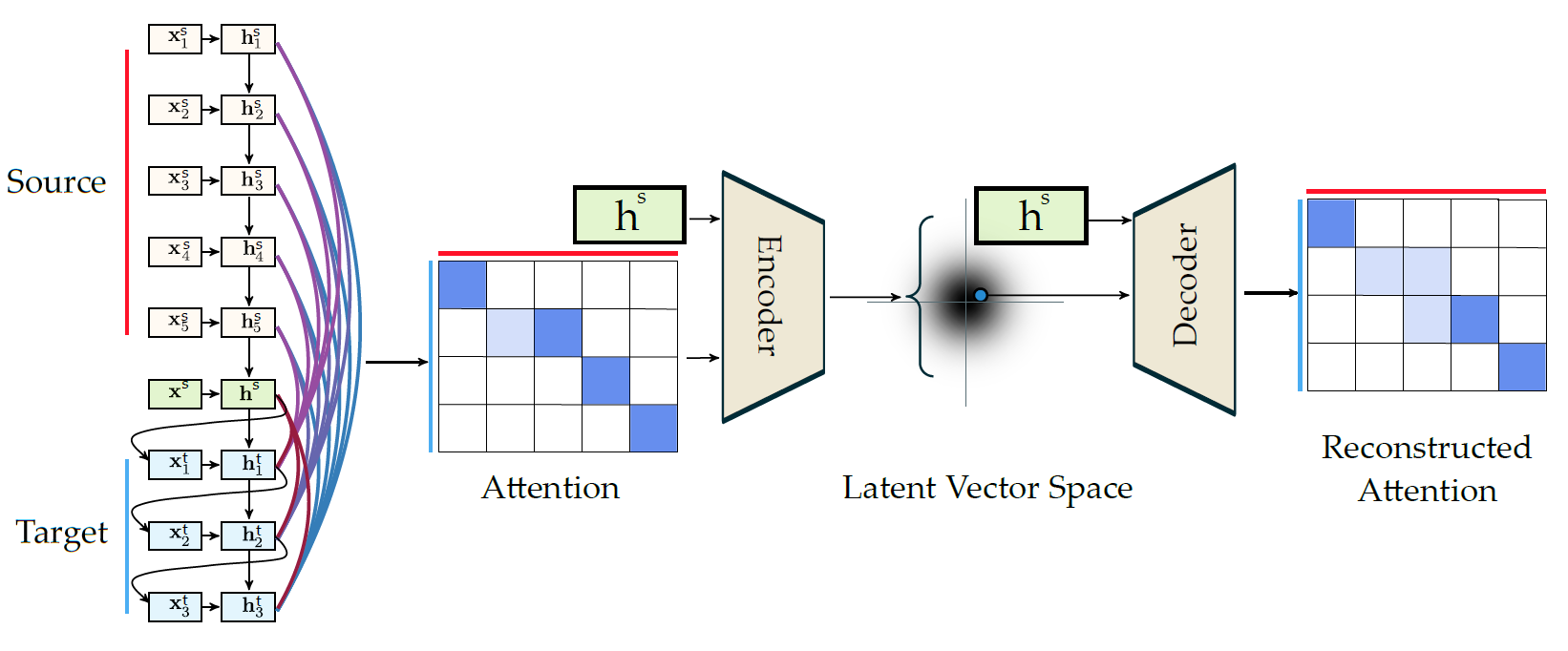}

\caption{Training of a conditional variational autoencoder applied to attention matrices. The seq2seq model translates training sentences from the source to a target domain while generating attention matrices. These matrices are concatenated with a representation of the source sentence and  encoded to a low dimensional latent vector space. 
}
\label{fig:att}
\end{figure*}
A sequence of input tokens is encoded to a series of hidden states using an encoder network and decoded to a target domain by a decoder network. 
During decoding, an attention mechanism is used to indicate which are the relevant input tokens at each step. This attention component is computed as an intermediate part of the model, and is trained jointly with the rest of the model.
Alongside being crucial for effective translation, attention --- while not necessarily correlated with human attention --- brings interpretability to seq2seq models by visualizing how individual input elements contribute to the model's decisions.
Attention values typically match up well with word alignments used in traditional statistical machine translation, obtained with tools such as GIZA++~\cite{OchP00-1056} or fast-align~\cite{fastN13-1073}.
Therefore, several works have included prior alignments from dedicated alignment software such as GIZA++ or fast-align \cite{AlignW16-2206,emnlp_suppattn,coling_supattn}. 
In particular, \citet{emnlp_suppattn} showed that the distance between the attention-infused alignments and the ones learned by an independent alignment model can be added to the networks' training objective, resulting in improved translation and alignment quality. 
Further, \citet{NIPS2017_7131} demonstrated that this alignment between given input sentence and generated output can be planned ahead as part of a seq2seq model: 
their model makes a plan of future alignments using an alignment-plan matrix and decides when to follow this plan by learning a separate commitment vector. 
In the standard seq2seq model, where attention is calculated at each time step, such overall alignment or focus is only apparent after decoding and is thus not carefully planned nor controlled.
We hypothesize that many text-to-text operations have varying levels of alignment and focus. 
To enable control over these aspects, we propose to pre-compute alignments and use this \textit{prior} attention to determine the structure or focus before decoding in order to steer output towards having specific attributes, such as length or level of compression. 
We facilitate this control through an input represented in a latent vector space (rather than, \eg explicit `style' attributes). 

After training of the initial seq2seq model (with standard attention) on a parallel text corpus, a conditional variational autoencoder~\cite{NIPS2015_5775} learns to reconstruct matrices of alignment scores or attention matrices from a latent vector space and the input sentence encoding.
At translation time, we are able to efficiently generate specific attention by sampling from regions in the latent vector space, resulting in output having specific stylistic attributes.  
We apply this method on a sentence simplification corpus, showing that we can control length and compression of output while producing realistic output and allowing fine-tuning for optimal evaluation scores.

\section{Generation of Prior Attention}
\label{sec:seq2seq}
This section describes our proposed method, sketched in \Figref{fig:att}, with emphasis on the generation of prior attention matrices.

An encoder recurrent neural network computes a sequence of representations over the source sequence, \ie its hidden states $\mathbf{h}^s_i$ (with $i=1,\ldots,n$ and $n$ the length of the source sequence). 
In attention-based models, an alignment vector $\mathbf{a}_j=[\alpha_{j,1},...,\alpha_{j,n}]$ 
is obtained by comparing the current target hidden state $\mathbf{h}^t_j$ with each source hidden state $\mathbf{h}^s_i$. 
A global context vector $\mathbf{c}_j$ is then computed as the weighted average, according to alignment weights of $\mathbf{a}_j$, over all the source states $\mathbf{h}^s_i$ at time step $j$ (for $j=1,...,m$ over $m$ decoding steps).
After decoding, these alignment vectors form a matrix $\mathbf{A}$ of attention vectors, $\mathbf{A} = [\mathbf{a}_1;\mathbf{a}_2;\ldots;\mathbf{a}_{m}]$,
capturing the alignment between source and target sequence.

\begin{table*}[h]
\footnotesize
	\centering
  \begin{tabular}{@{}ccp{14.5cm}@{}}
 \toprule
 $z_1$& $z_2$& \textbf{The wave traveled across the Atlantic , and organized into a tropical depression off the northern coast of Haiti on September 13 .} \\
  \textbf{--}   & \textbf{--} &   The wave traveled across the Atlantic , and organized into a tropical depression off the northern coast of the country on September 13 .\\
 \textbf{--}  & \textbf{+}   &  The wave traveled across the Atlantic Ocean into a tropical depression off the northern coast of Haiti on September 13 .\\
\textbf{+}   &  \textbf{--}  &  The wave traveled across the Atlantic Ocean and the Pacific Ocean to the south , and the Pacific Ocean to the south , and the Atlantic Ocean to the west .\\
\textbf{+}  &  \textbf{+}   &  The storm was the second largest in the Atlantic Ocean .\\
\midrule

 $z_1$& $z_2$& \textbf{Below are some useful links to facilitate your involvement .}\\

 \textbf{--}   & \textbf{--} &    Below are some useful links to facilitate your involvement .\\
  \textbf{--}  & \textbf{+}   &   Below are some useful links to help your involvement .\\
\textbf{+}   &  \textbf{--}  &  Below are some useful to be able to help help develop to help develop .\\
\textbf{+}  &  \textbf{+}   &   Below is a software program that is used to talk about what is now .\\
\bottomrule
\end{tabular}
\caption{Output excerpts for prior attention matrices sampled from a 2D latent vector space. Samples are drawn from outer regions, with $+$ indicating large positive values and $-$ for negative values.}
\label{tab:output_examples}
\end{table*}

Inspired by the field of image generation, we treat alignment matrices as grayscale images and use generative models to create previously unseen attention. 
Generative models have been applied to a variety of problems giving state-of-the-art results in image generation, text-to-speech synthesis, and image captioning. 
One of the most prominent models is the variational autoencoder (VAE) proposed by 
\citet{DBLP:journals/corr/KingmaW13}.
Given an observed variable $\mathbf{x}$, the VAE introduces a continuous latent variable $\mathbf{z}$, and assumes $\mathbf{x}$ to be generated from $\mathbf{z}$, i.e.,
$p(\mathbf{x}, \mathbf{z}) = p(\mathbf{x}|\mathbf{z}) \; p(\mathbf{z})$%
, with $p(\mathbf{z})$ being a prior over the latent variables.
%
$p_D(\mathbf{x}|\mathbf{z})$ is the conditional distribution that models the generation procedure parameterized by a decoder network $D$. For a given $\mathbf{x}$, an encoder network $E$ outputs a variational approximation $q_E(\mathbf{z}|\mathbf{x})$ of the true posterior over the latent values $p(\mathbf{z}|\mathbf{x}) \propto p_D(\mathbf{x}|\mathbf{z}) \; p_Z(\mathbf{z})$.
The parameters of ${E}$, ${D}$ are learned using stochastic variational inference to maximize a lower bound for the marginal likelihood of each observation in the training data. 
In our setting, $\mathbf{x}$ represents the attention matrix.

Next to control over stylistic features, we want attention matrices to be relevant for a specific source sentence. In the  Conditional Variational Autoencoder (CVAE)~\cite{conf/eccv/YanYSL16,NIPS2015_5775}, the standard VAE is conditioned on additional variables which can be used to generate diverse images conditioned on certain attributes, \eg generating different human faces given a sentiment.
We view the source contexts as the added conditional attributes and use the CVAE to generate diverse attention matrices instead of images.
%
This context vector is represented by the source sentence encoding $\mathbf{h}^s$. The CVAE encoder is conditioned on two variables, the attention matrix $\mathbf{A}$ and the sentence encoding $q_E(\mathbf{z}|\mathbf{A},\mathbf{h}^s)$. Analogously, for the decoder, the likelihood is now conditioned on two variables, a latent code $\mathbf{z}$ and again the source sentence encoding, $p_D(\mathbf{A}|\mathbf{z},\mathbf{h}^s)$.
%
This training procedure of the CVAE is visualized in \Figref{fig:att}.
At test time, the attention scores from the attention matrix, pre-generated from a latent code sample and the source sentence encoding, are used instead of the standard seq2seq model's attention mechanism. 
\begin{figure*}%
\centering
\begin{subfigure}[T]{0.465\textwidth}
\includegraphics[width=0.93\textwidth]{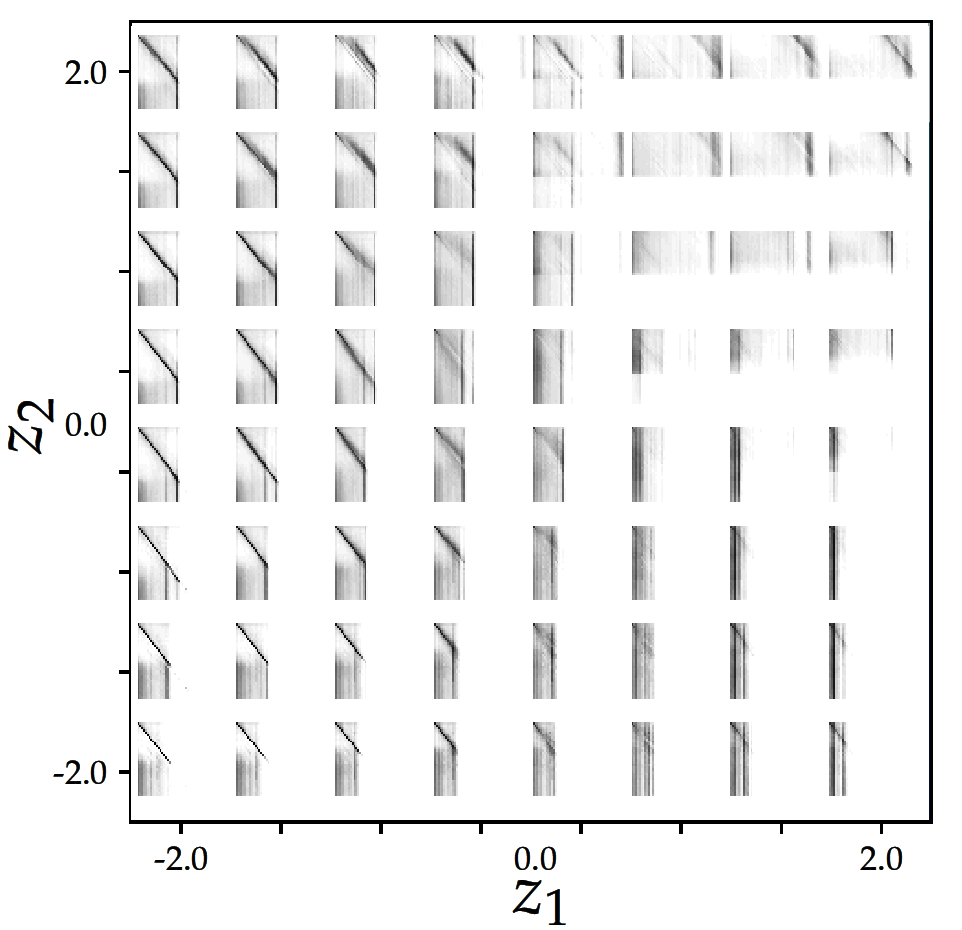}
\caption{}
\label{fig:zspace}
\end{subfigure}
\begin{subfigure}[T]{0.50\textwidth}%
\begin{center}
\includegraphics[width=.460\textwidth]{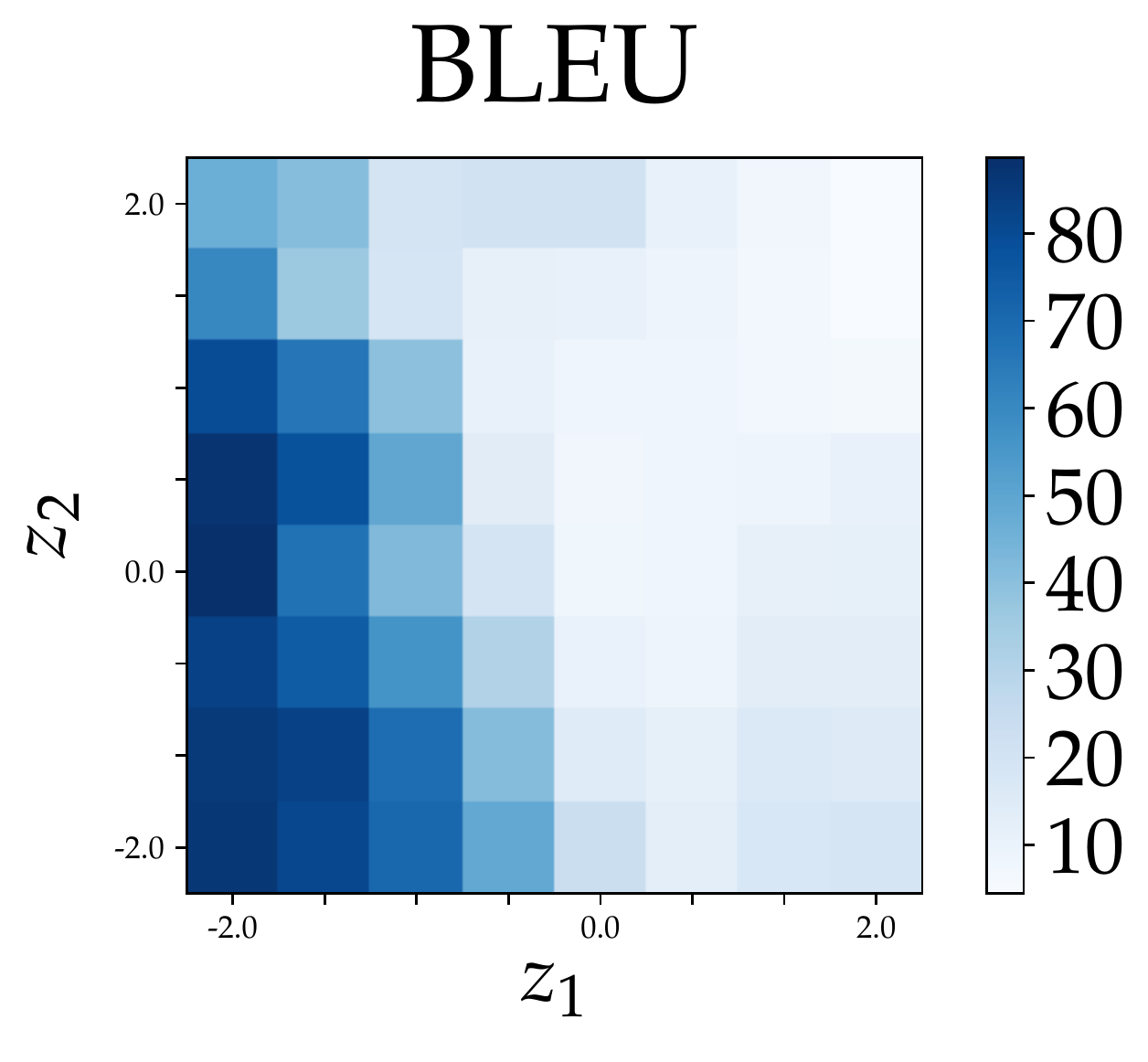}
\includegraphics[width=.460\textwidth]{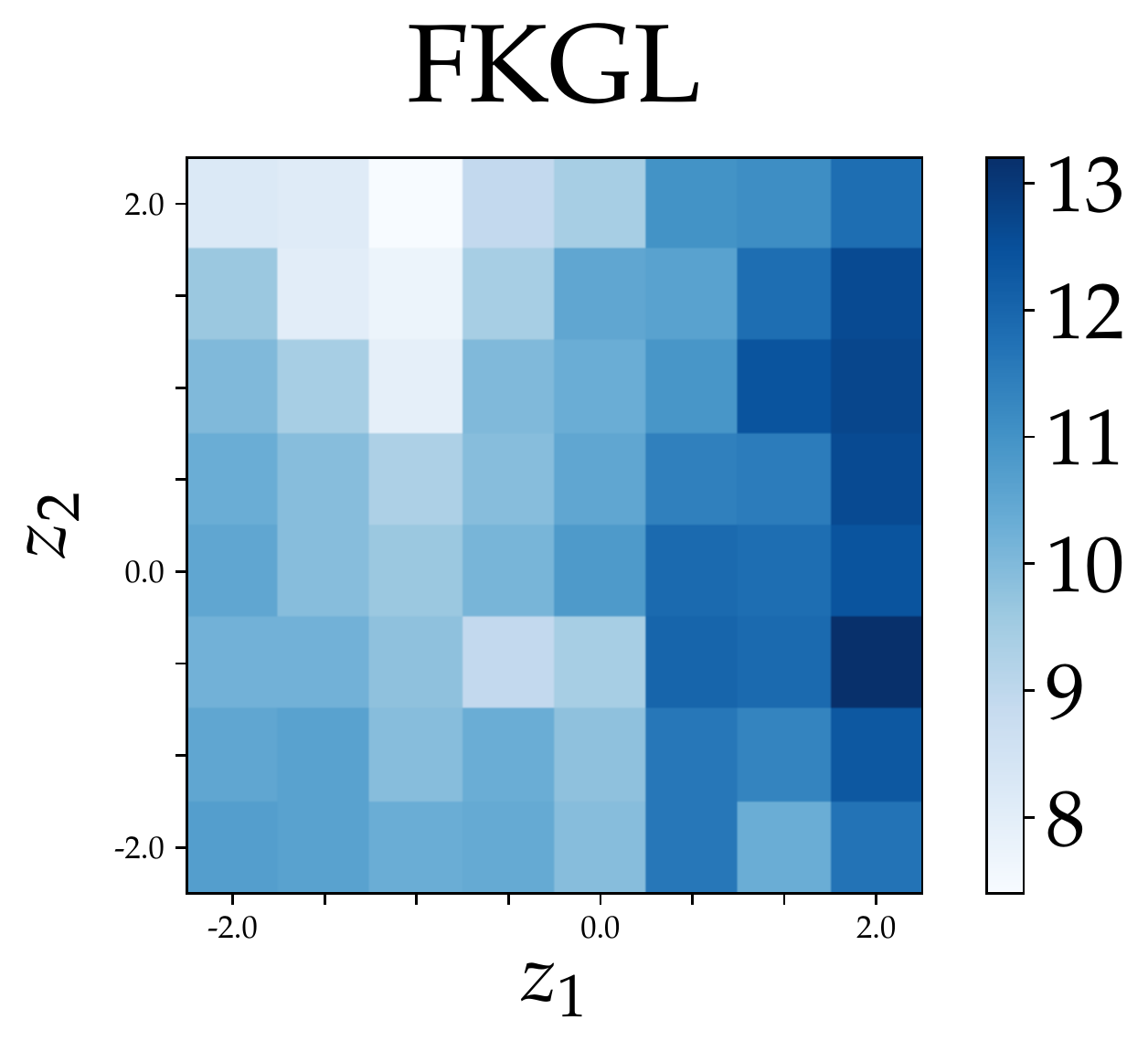}
\includegraphics[width=.460\textwidth]{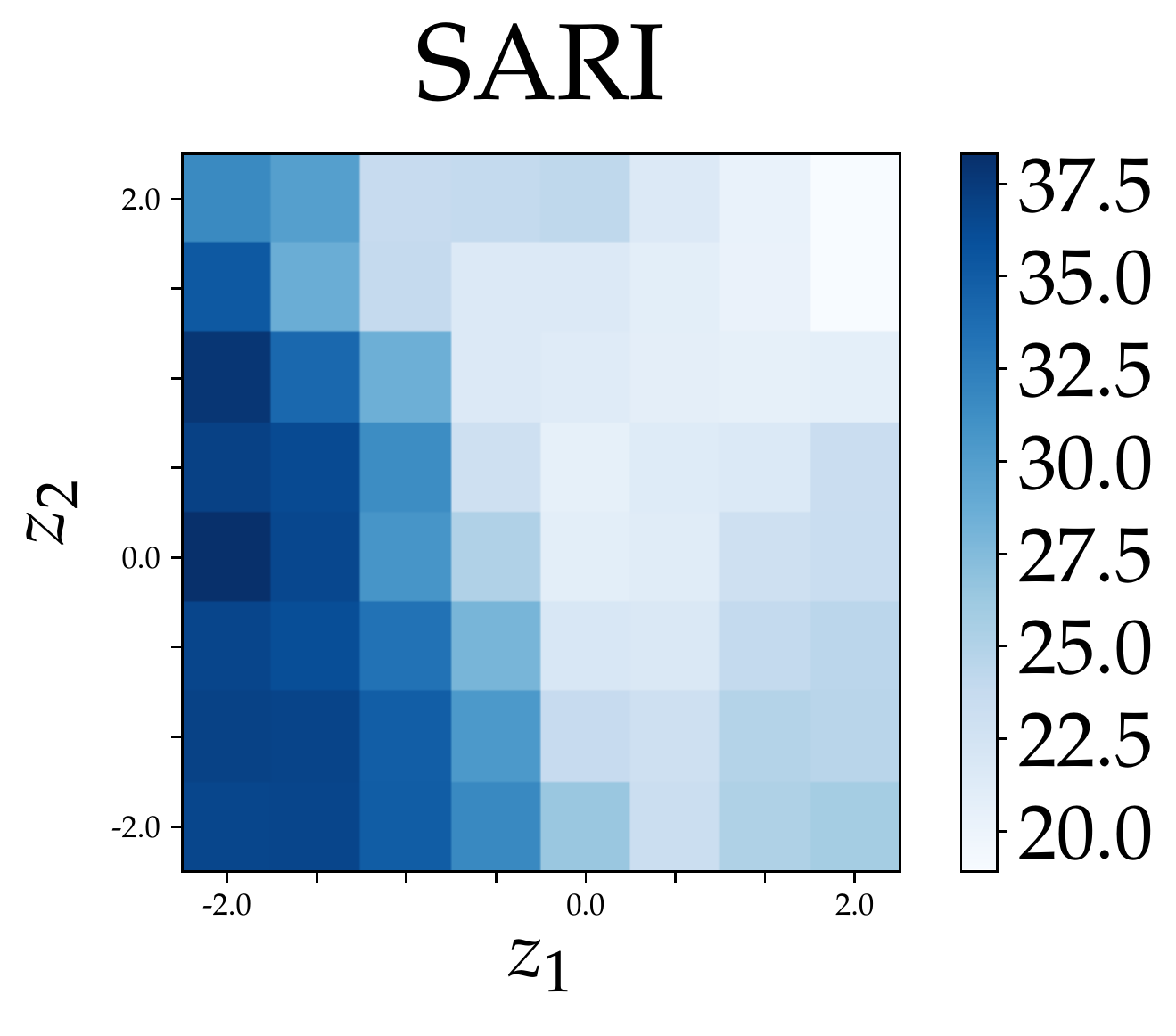}
\includegraphics[width=.460\textwidth]{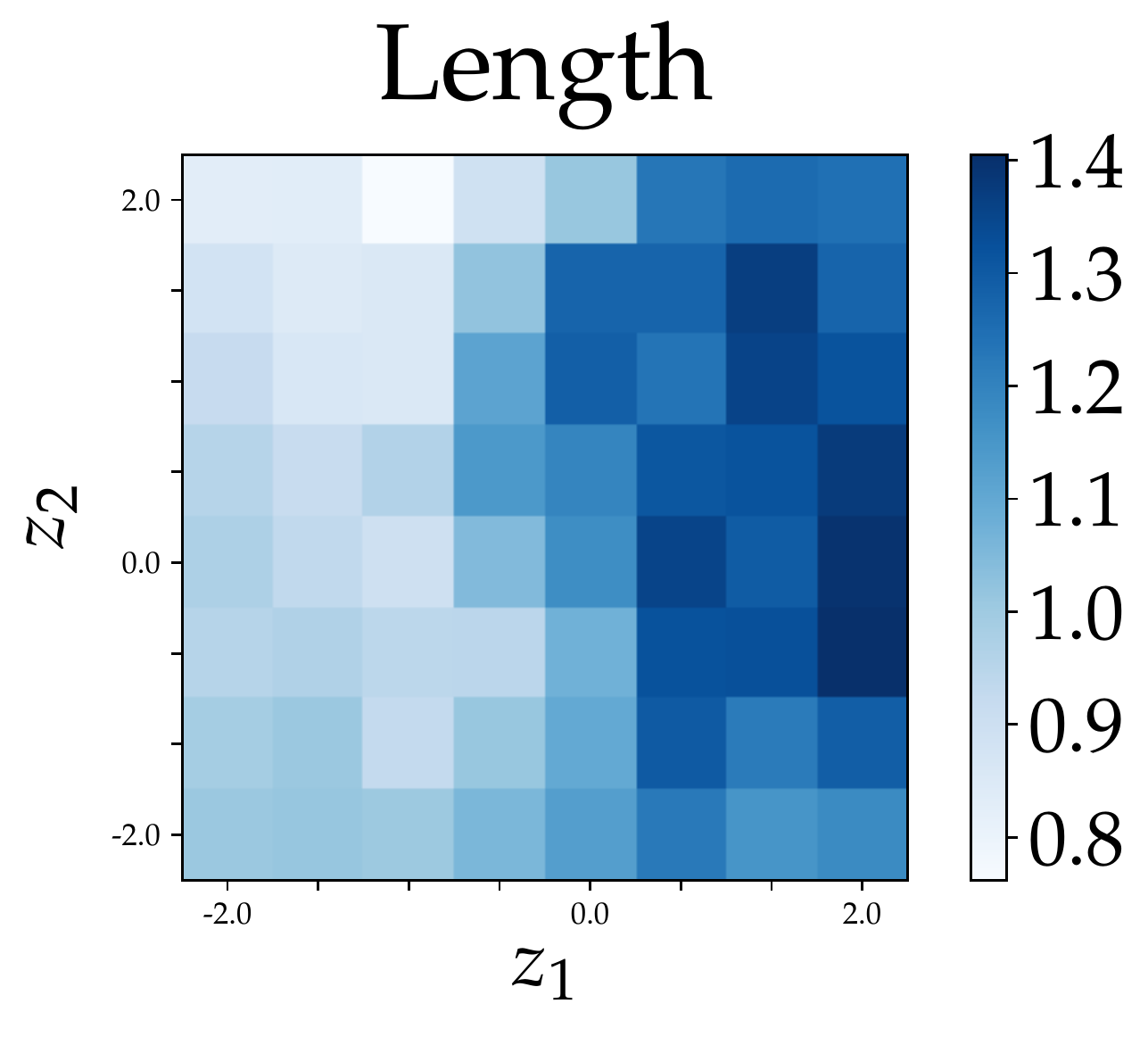}
\end{center}
\caption{}
\label{fig:scores}
\end{subfigure}
\caption{ (a) Attention matrices for a single source sentence encoding and a two-dimensional latent vector space. By conditioning the autoencoder on the source sentence, the decoder recognizes the length of the source and reduces attention beyond the last source token. (b) Score distributions for different regions of the latent vector space. }\label{fig:results}
\end{figure*}


\section{Experiments}
\subsection{Prior Attention for Text Simplification}
While our model is essentially task-agnostic, we demonstrate prior attention for the task of sentence simplification. The goal of sentence simplification is to reduce the linguistic complexity of text, while still retaining its original information and meaning. 
It has been suggested that sentence simplification can be defined by three major types of operations: \textit{splitting}, \textit{deletion}, and  \textit{paraphrasing}~\cite{Shardlow2014}.
We hypothesize that these operations occur at varying frequencies in the training data. 
We adopt our model in an attempt to capture these operations into attention matrices and the latent vector space, and thus control the form and degree of simplification through sampling from that space.
We train on the \textit{Wikilarge} collection used by Zhu~\shortcite{zhuC10-1152}.  
\textit{Wikilarge}
is a collection of 296,402 automatically aligned complex and simple sentences from the ordinary and simple English Wikipedia corpora, used extensively in previous work~\cite{DBLP:conf/acl/WubbenBK12,WoodsendD11-1038,ZhangD17-1062,P17-2014}. 
The training data includes 2,000 development and 359 test instances created by 
\citet{DBLP:journals/tacl/XuNPCC16}. These are complex sentences paired with simplifications provided by Amazon Mechanical Turk workers and provide a more reliable evaluation of the task. 
%
%

\begin{table}[h]
	\centering
    \small
  \begin{tabular}{@{}lcccc@{}}
  \toprule
& {BLEU} & {SARI}   &   {Length} & {FKGL}\\
\midrule
\cite{DBLP:conf/acl/WubbenBK12} & 67.74 & 35.34  & 0.90 & 10.0   \\
\cite{ZhangD17-1062}  & 90.00 & 37.62  & 0.95 & 10.4   \\
\cite{P17-2014} & 88.16 & 33.86  & 0.91 & 10.1   \\
\midrule
Seq2seq + online attention & 89.92 & 33.06  & 0.91 & 10.3   \\
Seq2seq + CVAE & 90.14 & 38.30  & 0.97 &  10.5  \\
\bottomrule
\end{tabular}
\caption{Quantitative evaluation of existing baselines from previous work and seq2seq with prior attention from the CVAE when choosing an optimal $z$ sample for BLEU scores. 
}
\label{tab:scores_standard}
\end{table}

\subsection{Hyperparameters and Optimization}
We extend the OpenNMT~\cite{2017opennmt} framework with functions for attention generation and release our code as a submodule.
%
We use a similar architecture as~\citet{zhuC10-1152} and \citet{P17-2014}: 2 layers of stacked unidirectional LSTMs with bi-linear global attention as proposed by \citet{DBLP:journals/corr/LuongLSVK15}, with hidden states of 512 dimensions.
The vocabulary is reduced to the 50,000 most frequent tokens and embedded in a shared 500-dimensional space.
We train using SGD with batches of 64 samples for 13 epochs after which the autoencoder is trained by translating sequences from training data.
%
Both the encoder and decoder of the CVAE comprise 2 fully connected layers of 128 nodes. Weights are optimized using ADAM~\cite{Kingma2014AdamAM}. 
We visualize and evaluate using a two-dimensional latent vector space. 
Source and target sequences are both padded or reduced to 50 tokens. 
The integration of the CVAE is analogous across the family of attention-based seq2seq models, \ie our approach can be applied more generally with different models or training data.
%
\subsection{Discussion}
To study the influence of sampling from different regions in the latent vector space, we visualize the resulting attention matrices and measure simplification quality using automated metrics.
\Figref{fig:zspace} shows the two-dimensional latent space for a single source sentence encoding using 64 samples ranging from values $-2$ to $2$. 
Next to the target-to-source length ratio, we apply automated measures commonly used to evaluate simplification systems~\cite{WoodsendD11-1038,ZhangD17-1062}:  BLEU, SARI~\cite{DBLP:journals/tacl/XuNPCC16},  FKGL\footnote{Fleish-Kincaid Grade Level index.}~\cite{kincaid1975derivation}.
Automated evaluation metrics for matrices originating from samples from different regions of latent codes are shown in \Figref{fig:scores}.
Inclusion of an attention mechanism was instrumental to match existing baselines. Our standard seq2seq model with attention, without prior attention, obtains a score of 89.92 BLEU points, which is close to scores obtained by similar models used in existing work on neural text simplification~\cite{ZhangD17-1062,P17-2014}.
In \tabref{tab:scores_standard}, we compare our seq2seq model with attention and without prior attention. 
A value for BLEU of 90.14 is found for $z=[-2,0]$ which was tuned on a development set. For the same $z$ value, a SARI value of 38.30 was reached. 
For comparison, we include the SMT-based model by \cite{DBLP:conf/acl/WubbenBK12}, the \textit{NTS} model by \cite{P17-2014} and the \textit{EncDecA} by \cite{ZhangD17-1062}.
For decreasing values of the first hidden dimension $z_1$, we observe that attention becomes situated at the diagonal, thus keeping closer to the structure of the source sentence and having one-to-one word alignments.
For increasing values of $z_1$, attention becomes more vertical and focused on single encoder states. This type of attention gives more control to the language model, as exemplified by output samples shown in \tabref{tab:output_examples}. Output from this region is far longer and less related to the source sentence. 

Influence of the second latent variable $z_2$ is less apparent from the attention matrices. However, sampling across this dimension shows large effects on evaluation metrics. For decreasing values, output becomes more similar to the source, with higher BLEU as a result. Sampling these values along the zero-axis results in the overall highest BLEU and SARI scores, trading similarity for simplification and readability.
%
%

\section{Conclusion}
We introduced a method to control the decoding process in sequence-to-sequence models using attention, in terms of stylistic characteristics of the output.
This means that the trained model is able to produce output with custom stylistic properties, given a well-chosen style input vector by the user at prediction time. 
Given the input sequence and an additional code vector to influence decoding characteristics, a variational autoencoder generates an attention matrix, which is used by the decoder to generate the output sequence according to the alignment style directed by the code vector.
We demonstrated the resulting variations in output for the task of text simplification. Yet, our method can be applied to any form of parallel text: we expect different types of training collections, such as translation or style transfer, to give rise to different characteristics or mappings in the latent space.


\bibliography{emnlp2018}
\bibliographystyle{acl_natbib_nourl}


\end{document}